%% file: main.tex
\pdfoutput=1

\documentclass[11pt]{article}

\usepackage[preprint]{acl}

\usepackage{times}
\usepackage{latexsym}

\usepackage{algorithm}

\usepackage{graphicx}%
\usepackage{multirow}%
\usepackage{amsmath,amssymb,amsfonts}%

\usepackage{amsthm}%
\usepackage{mathrsfs}%
\usepackage[title]{appendix}%
\usepackage{xcolor}%
\usepackage{textcomp}%
\usepackage{manyfoot}%
\usepackage{booktabs}%
\usepackage{algpseudocode}%
\usepackage{listings}%
\usepackage{anyfontsize}
\usepackage{color}
\usepackage{bbding}
\usepackage{longtable}
\usepackage{supertabular}
\usepackage{colortbl}

\usepackage[T1]{fontenc}

\usepackage[utf8]{inputenc}

\usepackage{microtype}

\usepackage{inconsolata}

\usepackage{graphicx}

\title{NAN: A Training-Free Solution to Coefficient Estimation in Model Merging}

\author{Chongjie Si$^1$, Kangtao Lv$^2$, Jingjing Jiang$^1$, Yadao Wang$^3$, Yongwei Wang$^2$,\\ \textbf{Xiaokang Yang$^1$, Wenbo Su$^3$, Bo Zheng$^3$, Wei Shen$^1$} \\
$^1$Shanghai Jiao Tong University, $^2$Zhejiang University, $^3$Alibaba Group\\
  \texttt{\{chongjiesi, wei.shen\}@sjtu.edu.cn}}

\begin{document}
\maketitle

\input{sec/0.abstract}
\input{sec/1.intro}

\input{sec/2.related}
\input{sec/3.method}
\input{sec/4.exp}
\input{sec/5.conclusion}

\bibliography{custom}
\appendix



\end{document}

%% file: sec/0.abstract.tex
\begin{abstract}
Model merging offers a training-free alternative to multi-task learning by combining independently fine-tuned models into a unified one without access to raw data. 
However, existing approaches often rely on heuristics to determine the merging coefficients, limiting their scalability and generality.
In this work, we revisit model merging through the lens of least-squares optimization and show that the optimal merging weights should scale with the amount of task-specific information encoded in each model. 
Based on this insight, we propose \textbf{NAN}, a simple yet effective method that estimates model merging coefficients via the inverse of parameter norm. 
NAN is training-free, plug-and-play, and applicable to a wide range of merging strategies. 
Extensive experiments on show that NAN consistently improves performance of baseline methods.
\end{abstract}

%% file: sec/1.intro.tex
\section{Introduction}
The widespread adoption of pre-trained models (PTMs) has revolutionized both NLP and CV by enabling efficient task-specific fine-tuning with minimal annotated data \cite{devlin2019bert,dosovitskiy2020image,raffel2020exploring}. Public model hubs such as HuggingFace Transformers \cite{wolf2020transformers}, timm, and torchvision have accelerated the release of numerous backbone and fine-tuned checkpoints, leading to a rapid proliferation of task-specialized models. However, maintaining a separate model for each task imposes substantial storage and deployment overhead, posing scalability challenges in multi-task scenarios \cite{ruder2016overview}. While multi-task learning (MTL) offers a potential solution by jointly training on multiple tasks \cite{caruana1997multitask}, it is hindered by high computational costs, the need for simultaneous access to all datasets, and complexities in balancing heterogeneous task objectives \cite{jin2022dataless}.

To address the limitations of task-specific fine-tuning and the overhead of multi-task training, model merging has emerged as a promising paradigm for integrating independently fine-tuned models without access to training data \cite{ilharco2022editing, kinderman2024foldable, yadav2023ties}.
While naive weight averaging often fails due to parameter misalignment \cite{wortsman2022model}, recent works have proposed more principled approaches involving importance weighting, task-vector manipulation, and pre-processing techniques. 
These methods demonstrate that, with appropriate alignment and weighting, model merging can serve as an efficient and modular alternative to multi-task learning.
Despite the promising progress, model merging still faces a fundamental challenge: many existing methods rely on heuristic or intuitive strategies for weight combination coefficients \cite{yadav2023ties, ilharco2022editing}, lacking rigorous theoretical justification. 
These limitations prompt a reconsideration of how model merging should be fundamentally approached.

In this work, we revisit the fundamental principles of model merging and propose a theoretically grounded framework.
Starting from a least-squares formulation, we derive the optimal merging coefficients and reveal that the ideal merging weights should be proportional to the amount of task-specific information encoded in each model. 
Building on this insight, we introduce NAN, a novel training-free model merging plugin that leverages this information-theoretic perspective to achieve effective integration of multiple fine-tuned models. 
Extensive experiments demonstrate the effectiveness and generality of our approach, with NAN improving the performances of baseline methods.

%% file: sec/2.related.tex
\section{Related Work}
Model merging aims to integrate multiple task-specific models into a single one, reducing the need to store and manage separate models for each task \cite{jin2022dataless,yadav2023ties,yang2023adamerging,stoica2023zipit,yu2024language,ilharco2022editing}.
While naive weight averaging \cite{wortsman2022model} is simple, it often leads to severe performance drops due to parameter misalignment. 
To overcome this, various methods estimate merging coefficients using heuristics or additional statistics. For instance, Fisher-Merging \cite{matena2022merging} and RegMean \cite{jin2022dataless} rely on Fisher or inner-product matrices, which must be provided or computed manually. 
Task vector-based approaches such as Task Arithmetic \cite{ilharco2022editing}, Ties-Merging \cite{yadav2023ties}, and AdaMerging \cite{yang2023adamerging} define merging in the space of model deltas, but their success heavily depends on intuitively selected or hand-tuned coefficients. 
Although AdaMerging estimates coefficients adaptively, it still assumes access to model-specific conditions. 
DARE \cite{yu2024language} sparsifies task vectors to reduce interference but shows limited gains and is only tested on a small number of tasks.
Overall, most existing methods require either auxiliary information or strong manual heuristics.

%% file: sec/3.method.tex
\section{Method}

In this section, we conduct an in-depth exploration of model merging from the perspective of least squares optimization.

\subsection{Model Merging via Least Squares}

To better understand the underlying principles of model merging, we begin with a simplified least-squares formulation. 
Suppose we have two tasks, each associated with data matrices $\mathbf{X}_1 \in \mathbb{R}^{n_1 \times d}$, $\mathbf{Y}_1 \in \mathbb{R}^{n_1 \times m}$, and $\mathbf{X}_2 \in \mathbb{R}^{n_2 \times d}$, $\mathbf{Y}_2 \in \mathbb{R}^{n_2 \times m}$, where $\mathbf{X}_i$ represents input features and $\mathbf{Y}_i$ denotes task-specific supervision.
For each task, we consider an independent least-square problem as:
\begin{equation}
\begin{aligned}
     & \min_{\mathbf{W}_1} \|\mathbf{X}_1 \mathbf{W}_1 - \mathbf{Y}_1\|_F^2 \\ 
    & \min_{\mathbf{W}_2} \|\mathbf{X}_2 \mathbf{W}_2 - \mathbf{Y}_2\|_F^2,
\end{aligned}
\end{equation}
whose solutions admit closed forms:
\begin{equation}
\begin{aligned}
    & \mathbf{W}_1^* = (\mathbf{X}_1^\mathsf{T} \mathbf{X}_1)^{-1} \mathbf{X}_1^\mathsf{T} \mathbf{Y}_1, \\
    & \mathbf{W}_2^* = (\mathbf{X}_2^\mathsf{T} \mathbf{X}_2)^{-1} \mathbf{X}_2^\mathsf{T} \mathbf{Y}_2. 
\end{aligned}   
\end{equation}
Now consider the joint least-squares objective that seeks a shared model $\mathbf{W}$ across both tasks:
\begin{equation}
    \min_{\mathbf{W}} \|\mathbf{X}_1 \mathbf{W} - \mathbf{Y}_1\|_F^2 + \|\mathbf{X}_2 \mathbf{W} - \mathbf{Y}_2\|_F^2.
\end{equation}
This problem has the following closed-form solution:
$\mathbf{W}^* = (\mathbf{X}_1^\mathsf{T} \mathbf{X}_1 + \mathbf{X}_2^\mathsf{T} \mathbf{X}_2)^{-1} (\mathbf{X}_1^\mathsf{T} \mathbf{Y}_1 + \mathbf{X}_2^\mathsf{T} \mathbf{Y}_2)$.
To explore the relationship between the jointly optimized solution $\mathbf{W}^*$ and the individually optimized $\mathbf{W}_1^*$ and $\mathbf{W}_2^*$, we note that:
\begin{equation}
    \mathbf{W}_1^* = \mathbf{A}_1^{-1} \mathbf{b}_1, \quad \mathbf{W}_2^* = \mathbf{A}_2^{-1} \mathbf{b}_2,
\end{equation}
where $\mathbf{A}_i = \mathbf{X}_i^\mathsf{T} \mathbf{X}_i$, $\mathbf{b}_i = \mathbf{X}_i^\mathsf{T} \mathbf{Y}_i$. 
Then,
\begin{equation}
\mathbf{W}^* = (\mathbf{A}_1 + \mathbf{A}_2)^{-1} (\mathbf{b}_1 + \mathbf{b}_2).    
\end{equation}
We now attempt to express $\mathbf{W}^*$ as a weighted combination of $\mathbf{W}_1^*$ and $\mathbf{W}_2^*$. 
Observe that:
\begin{equation}
    \mathbf{W}^* = (\mathbf{A}_1 + \mathbf{A}_2)^{-1} \left(\mathbf{A}_1 \mathbf{W}_1^* + \mathbf{A}_2 \mathbf{W}_2^*\right),
\end{equation}
this leads to:
\begin{equation}
    \mathbf{W}^* = \boldsymbol{\Omega}_1 \mathbf{W}_1^* + \boldsymbol{\Omega}_2 \mathbf{W}_2^*,    
\end{equation}
where the merging coefficients are matrix-valued:
\begin{equation}
\begin{aligned}
     & \boldsymbol{\Omega}_1 = (\mathbf{A}_1 + \mathbf{A}_2)^{-1} \mathbf{A}_1 \\
    & \boldsymbol{\Omega}_2 = (\mathbf{A}_1 + \mathbf{A}_2)^{-1} \mathbf{A}_2.  
\end{aligned}
\end{equation}
This formulation reveals that the optimal merged solution is a weighted average of the individual solutions, where the weights are determined by the relative information content of each task, as quantified by $\mathbf{X}_i^\mathsf{T} \mathbf{X}_i$—essentially the unnormalized covariance matrix of the inputs. In other words, tasks with more informative or higher-variance input distributions contribute more to the merged solution.

\subsection{Sample-Weighted Merging}

To further understand the behavior of the merging coefficients, we now consider the case where the input features are normalized.
This is a common pre-processing step in deep learning pipelines, especially in representation learning and contrastive objectives.
Under this normalization, the matrix $\mathbf{A}_i = \mathbf{X}_i^\mathsf{T} \mathbf{X}_i$ becomes approximately proportional to the sample size $n_i$, assuming the features are approximately isotropic:
$\mathbf{A}_i \approx n_i \mathbf{I}_d$,
where $\mathbf{I}_d$ is the $d$-dimensional identity matrix. 
Substituting this into the earlier expression for the merged solution yields:
\begin{equation}
    \mathbf{W}^* \approx \frac{n_1 \mathbf{W}_1^* + n_2 \mathbf{W}_2^*}{n_1 + n_2},
\end{equation}
This result provides a simple yet powerful insight: under normalized input features, the optimal merged model is approximately a sample-size-weighted average of the individually fine-tuned models. 
Consequently, the relative contribution of each model should be proportional to the amount of data it was trained on.

\subsection{NAN: A Training-Free Plugin}

In practice, when the exact values of $n_1$ and $n_2$ are not available—such as when merging open-source fine-tuned models—the direct estimation of sample sizes becomes infeasible.
To address this, we resort to empirical proxies that reflect the amount of information each model has absorbed during fine-tuning.

Recent findings suggest that the variance of the learned weights is inversely correlated with the training data volume \cite{fort2019deep, izmailov2018averaging, si2025unveiling, du2025loca}, i.e., $n \propto \frac{1}{\operatorname{Var}(\mathbf{W})}$.
Intuitively, models trained on larger datasets exhibit lower variance in parameter updates, as the optimization process averages out stochastic fluctuations over more samples. 
This observation provides us with a practical prior for estimating task importance.
Given that most pre-trained and fine-tuned weights are approximately zero-centered \cite{du2025loca,si2025unveiling}, we adopt the variance of the weights as a proxy signal. 
Assuming zero-mean updates, we have: $\operatorname{Var}(\mathbf{W}) \propto \|\mathbf{W}\|_F^2$,
where the Frobenius norm serves as a direct measure of magnitude.
In practice, we adopt the Frobenius norm rather than its squared value to compute the merging coefficients, as the squared norm may introduce large scaling disparities and result in numerical instability during normalization. 
The norm itself offers a more stable approximation while still reflecting the relative importance of each model.

Combining this insight with our earlier derivation that optimal merging weights should scale with the sample size, we introduce \textbf{N}orm-\textbf{A}ware mergi\textbf{N}g (NAN), a training-free plug-in.
Specifically, given $m$ task-specific models to be merged, NAN computes the Frobenius norm of each model's weights $\mathbf{W}$ as:
\begin{equation}
\alpha_i = \frac{1 / \|\mathbf{W}_i\|_F}{\sum_{j=1}^m 1 / \| \mathbf{W}_j\|_F}.
\end{equation}
When merging a large number of models, the softmax-normalized coefficients can become excessively small.
To mitigate this issue, we apply a global scaling factor $m/2$ to the merged weights. 
NAN is highly versatile and can be seamlessly integrated into any existing model merging pipeline. 
It can be applied either directly on raw model weights or as a post-processing reweighting step following other merging strategies.

%% file: sec/4.exp.tex
\section{Experiment}

\textbf{Baselines.} We compare NAN against the following baselines: Individual Models, Traditional Multi-task Learning, the training-based method AdaMerging \cite{yadav2023ties}, and several training-free methods, including Weight Averaging \cite{wortsman2022model}, Fisher Merging \cite{matena2022merging}, RegMean \cite{jin2022dataless}, Task Arithmetic \cite{ilharco2022editing}, and Ties-Merging \cite{yadav2023ties}.

\vspace{1mm}
\noindent
\textbf{Vision Task.}
Following prior work \cite{yadav2023ties,yang2023adamerging}, we adopt ViT-B/32 and ViT-L/14 as the pre-trained backbone for all methods.
Evaluation is conducted across eight image classification tasks: SUN397 \cite{xiao2010sun}, Cars \cite{krause20133d}, RESISC45 \cite{cheng2017remote}, EuroSAT \cite{helber2019eurosat}, SVHN \cite{netzer2011reading}, GTSRB \cite{stallkamp2011german}, MNIST \cite{lecun1998mnist}, and DTD \cite{cimpoi2014describing}.
All datasets are evaluated using top-1 classification accuracy as the performance metric.

\input{sec/table/vit}

Table~\ref{tab:performance_vitbase32} shows the performance of various merging methods.
While individual models and multi-task learning provide strong baselines, training-based methods require additional optimization and metadata. 
Among training-free approaches, NAN achieves consistently better performance when coupling with baseline methods. 
This demonstrates NAN's effectiveness as a simple and general merging strategy without relying on task-specific tuning or training.

\vspace{1mm}
\noindent
\textbf{Language Task.}
Following prior work \cite{yu2024extend}, we use LLaMA2-13B \cite{touvron2023llama2} as the backbone and merge two of its fine-tuned variants: WizardLM-13B \cite{xu2024wizardlm} and WizardMath-13B \cite{luo2023wizardmath}.
We test the performance on four datasets: MMLU \cite{hendryckstest2021}, CEval \cite{huang2023ceval}, GSM8K \cite{cobbe2021gsm8k}, and BBH \cite{suzgun2022challenging}.
The results on GSM8K is evaluated following the official protocol of the Qwen2.5 Math Eval Toolkit \cite{yang2024qwen2}, while others are obtained using the OpenCompass evaluation framework \cite{2023opencompass}.

\begin{table}[ht]
 \renewcommand\arraystretch{1}
 \setlength{\tabcolsep}{1.3mm}
    \centering
    \caption{Results on language merging tasks.}
    \resizebox{\linewidth}{!}{
    \begin{tabular}{c | c c c c | c}
        \toprule
        Method & MMLU & CEval & GSM8K &  BBH & \textbf{Avg} \\ 
        \midrule
        WizardLM-13B & 53.6 & 32.6 & 38.8 & 19.4 & 36.1  \\
        WizardMath-13B & 54.2 & 37.7 & 46.9 & 44.8 & 45.9 \\
        
        \midrule         
        
        Task Arithmetic (TA) & 56.3 & 39.5 & 52.7 & 35.7 & 46.0  \\

        \rowcolor{gray!20}

        TA + NAN & 56.3 & 38.8 & 64.1 & 34.6 & \textbf{48.5} \\

        Ties-Merging (Ties) & 55.9 & 40.0 & 55.3 & 38.9 & 47.5 \\ 

        Ties + NAN & 56.8 & 39.2 & 58.5 & 39.3 & 48.5 \\
        
        \bottomrule
    \end{tabular}
    }
    \label{tab model merging llm}
\end{table}

Table \ref{tab model merging llm} shows the results of merging two LLaMA2-13B variants on four language understanding and reasoning benchmarks. 
Task Arithmetic and Ties-Merging both improve over the individual models, indicating the benefits of parameter fusion. Our method achieves further gains, particularly on GSM8K, and yields the highest average performance across all datasets, demonstrating its effectiveness in merging complementary capabilities from general-purpose and math-specialized models.

\vspace{1mm}
\noindent
\textbf{VLM Task.}
Following prior work \cite{si2025unveiling}, we adopt the vision-language model (VLM) LLaVA-v1.5-13B \cite{liu2023llava} as the shared pre-trained base model and merge two of its fine-tuned variants: LLaVA-v1.6-13B \cite{liu2023llava}, optimized for general multi-modal understanding, and Math-LLaVA \cite{shi2024math}, which is specialized for mathematical reasoning.
We test the performance on four datasets: MathVista \cite{lu2023mathvista}, WeMath \cite{qiao2024we}, AI2D \cite{kembhavi2016diagram}, and GeoQA \cite{chen2021geoqa}.

\begin{table}[ht]
 \renewcommand\arraystretch{1}
 \setlength{\tabcolsep}{1mm}
    \centering
    \caption{Results on VLM merging tasks.}
    \resizebox{\linewidth}{!}{
    \begin{tabular}{c | c c c c | c}
        \toprule
        Method & MathVista & WeMath & AI2D &  GeoQA & \textbf{Avg} \\ 
        \midrule
        LLaVA-v1.5-13B & 34.3 & - & 61.1 & - & - \\
        \midrule
        LLaVA-1.6-13B & 33.6 & 30.1 & 67.9 & 23.9 & 38.9 \\ 
       
        Math-LLaVA & 45.8 & 33.9 & 66.7 & 46.6 & 48.3 \\ 

         \midrule
         
        Task Arithmetic (TA) & 43.7 & 35.2 & 69.3 & 41.2 & 47.4 \\ 
        
        \rowcolor{gray!20}
        
        TA + NAN & 44.9 & 36.5 & 67.2 & 46.6 & \textbf{48.8} \\
        
        \bottomrule
    \end{tabular}
    }
    \label{tab model merging}
\end{table}

Table \ref{tab model merging} summarizes the results of merging two LLaVA-based models across four visual-language reasoning benchmarks.
Compared to the individual models, Task Arithmetic achieves a reasonable trade-off, but still underperforms the task-specialized Math-LLaVA on certain datasets.
By incorporating NAN into Task Arithmetic, we observe consistent improvements across most tasks, leading to the best overall average.
This demonstrates that NAN can effectively enhance existing merging strategies in the multi-modal setting.

%% file: sec/table/vit.tex
\begin{table*}[ht]
    \renewcommand\arraystretch{0.9}
    \centering
    \caption{Multi-task performance when merging ViT-B/32 and ViT-L/14 models on eight tasks.}
    \resizebox{\textwidth}{!}{  
    \begin{tabular}{l|cccccccc|c}
    \toprule
    
    Method & SUN397 & Cars & RESISC45 & EuroSAT & SVHN & GTSRB & MNIST & DTD & \textbf{Avg Acc} \\
    
    \midrule
    \multicolumn{10}{c}{{ViT-B/32}} \\
    \midrule   
    
    Pretrained & 62.3 & 59.7 & 60.7 & 45.5 & 31.4 & 32.6 & 48.5 & 43.8 & 48.0 \\
    
    Individual & 75.3 & 77.7 & 96.1 & 99.7 & 97.5 & 98.7 & 99.7 & 79.4 & 90.5 \\
    
    Traditional MTL & 73.9 & 74.4 & 93.9 & 98.2 & 95.8 & 98.9 & 99.5 & 77.9 & 88.9 \\ \midrule
    
    AdaMerging++ & 60.8 & 56.9 & 73.1 & 83.4 & 87.3 & 82.4 & 95.7 & 50.1 & 73.7 \\
    
    Layer-wise AdaMerging & 64.5 & 68.1 & 79.2 & 93.8 & 87.0 & 91.9 & 97.5  & 59.1 & 80.1 \\ \midrule

    Weight Averaging & 65.3 & 63.4 & 71.4 & 71.7 & 64.2 & 52.8 & 87.5 & 50.1 & 65.8 \\
    
    Fisher Merging & 68.6 & 69.2 & 70.7 & 66.4 & 72.9 & 51.1 & 87.9 & 59.9 & 68.3 \\
    
    RegMean & 65.3 & 63.5 & 75.6 & 78.6 & 78.1 & 67.4 & 93.7 & 52.0 & 71.8 \\
    
    Task Arithmetic (TA) & 55.2 & 54.9 & 66.7 & 78.9 & 80.2 & 69.7 & 97.3 & 50.4 & 69.1 \\

    \rowcolor{gray!20}

    TA + NAN & 59.3 & 58.2 & 69.7 & 83.3 & 76.2 & 71.0 & 96.1 & 61.6 & 70.7 \\
    
    Ties-Merging (Ties) & 59.8  & 58.6 & 70.7 & 79.7 & 86.2 & 72.1 & 98.3 & 54.2 & 72.4 \\

    \rowcolor{gray!20}

    Ties+NAN & 61.6 & 61.8 &  74.0 & 80.9 & 83.8 & 75.7 & 97.8 & 54.6 & \textbf{73.8} \\
    
    \midrule
    
    \multicolumn{10}{c}{{ViT-L/14}} \\
    \midrule  
    Individual & 82.3 & 92.4 & 97.4 & 100 & 98.1 & 99.2 & 99.7 & 84.1 & 94.2 \\
    
    Traditional MTL & 80.8 & 90.6 & 96.3 & 96.3 & 97.6 & 99.1 & 99.6 & 84.4 & 93.5 \\ \midrule

    Task Arithmetic & 74.1 & 82.1 & 86.7 & 93.8 & 87.9 & 86.8 & 98.9 & 65.6 & 84.5 \\
    
    Ties-Merging (Ties) & 76.5 & 85.0 & 89.3 & 95.7 & 90.3 & 83.3 & 99.0 & 68.8 & 86.0 \\

    \rowcolor{gray!20}

    Ties + NAN & 74.4 & 84.3 & 87.7 & 95.3 & 89.5 & 92.5 & 99.2 & 68.5 & \textbf{86.4} \\
    
    \bottomrule
    \end{tabular}
    }
 \label{tab:performance_vitbase32} 
\end{table*}

%% file: sec/5.conclusion.tex
\section{Conclusion}

In this work, we present NAN, a novel training-free model merging framework grounded in a principled least-squares formulation. 
By interpreting model merging through the lens of theory, we derive theoretically optimal merging coefficients that reflect the task-specific knowledge embedded in each fine-tuned model. 
This perspective enables a simple yet effective merging plugin that circumvents the computational burden and retraining requirements of traditional multi-task learning or heuristic-based merging approaches. 
Our extensive empirical evaluation confirms the generality and robustness of NAN, consistently achieving competitive or superior performance compared to existing baselines.

\section*{Limitations}
While NAN demonstrates strong performance across various domains, it currently focuses on merging models with a shared pre-trained backbone and may require adaptation for merging across heterogeneous architectures or modalities.